\documentclass[
]{ceurart}

\usepackage{listings}
\usepackage{cite}
\usepackage{url} 
\usepackage{amsmath,amssymb,amsfonts}
\usepackage{algorithm}
\usepackage{natbib} 
\usepackage{graphicx}
\usepackage{textcomp}
\usepackage{xcolor}
\usepackage{adjustbox}
\usepackage{booktabs}
\usepackage{algpseudocode}
\usepackage{enumitem}
\usepackage{natbib}
\usepackage{graphicx}
\usepackage{booktabs}
\usepackage{multirow}
\usepackage{amsmath}
\usepackage{float}
\begin{document}

\copyrightyear{2025}
\copyrightclause{Copyright for this paper by its authors.
  Use permitted under Creative Commons License Attribution 4.0
  International (CC BY 4.0).}

\conference{TRUST-AI: The European Workshop on Trustworthy AI. Organized as part of the European Conference of Artificial Intelligence - ECAI 2025. October 2025, Bologna, Italy.}

\title{Diverse And Private Synthetic Datasets Generation for RAG evaluation: A multi-agent framework}

\author[1]{Ilias DRIOUICH}[email=ilias.driouich@amadeus.com]

\author[1]{Hongliu CAO}

\author[1]{Eoin THOMAS}
\address[1]{AMADEUS France}

\cortext[1]{Corresponding author.}


\begin{abstract}
Retrieval-augmented generation (RAG) systems improve large language model outputs by incorporating external knowledge, enabling more informed and context-aware responses. However, the effectiveness and trustworthiness of these systems critically depends on how they are evaluated, particularly on whether the evaluation process captures real-world constraints like protecting sensitive information. While current evaluation efforts for RAG systems have primarily focused on the development of performance metrics, far less attention has been given to the design and quality of the underlying evaluation datasets, despite their pivotal role in enabling meaningful, reliable assessments. In this work, we introduce a novel multi-agent framework for generating synthetic QA datasets for RAG evaluation  that prioritize semantic diversity and privacy preservation. Our approach involves: (1) a Diversity agent leveraging clustering techniques to maximize topical coverage and semantic variability, (2) a Privacy Agent that detects and mask sensitive information across multiple domains and (3) a QA curation agent that synthesizes private and diverse QA pairs suitable as ground truth for RAG evaluation. Extensive experiments demonstrate that our evaluation sets outperform baseline methods in diversity and achieve robust privacy masking on domain-specific datasets. This work offers a practical and ethically aligned pathway toward safer, more comprehensive RAG system evaluation, laying the foundation for future enhancements aligned with evolving AI regulations and compliance standards.

\end{abstract}



\maketitle

\section{Introduction}


Retrieval-augmented generation (RAG) aims to improve large language models (LLM) output by incorporating relevant information retrieved from external knowledge sources. It has been effectively applied in various scenarios, such as domain-specific chatbots \citep{siriwardhana2023improving, cao2024recent} and email/code completion \citep{parvez2021retrieval}. A typical RAG system often operates in two stages: retrieval and generation. First, the system retrieves the relevant knowledge from an external database based on the user query. Then, the retrieved information is integrated with the query to form an input for the LLM in charge of the generation stage. The LLM uses its pre-trained knowledge and the retrieval data to generate a response, enhancing the overall quality of the output. As RAG sees wider adoption, ensuring robust performance evaluation becomes critical. While numerous automated methods, ranging from classic n-gram metrics (BLEU, ROUGE) and embedding-based measures (BERTScore)\citep{scores} to the “LLM-as-a-judge”  approach  leveraging GPT have been explored, an equally vital element is having the “golden” evaluation set with sufficiently diverse and representative samples that will serve as a complete benchmark to evaluate both the retrieval and generation processes \citep{cao2025multi}. Furthermore, despite the emergence of multiple RAG benchmarks \citep{joshi2017triviaqa, chen2024benchmarking, lyu2024crud} that span general-purpose and specialized domains, many still fall short of reflecting the complexity and variability of real-world use cases. In particular, traditional benchmarks often lack coverage of novel or underrepresented topics, limiting their ability to generalize \citep{cao2024recent, cao2025enhancing}. This gap poses a significant challenge for reliable evaluation, especially in domains requiring deep expertise and factual precision \citep{bruckhaus2024ragdoesworkenterprises}.

To address these challenges, a new line of work consisting in generating synthetic evaluation sets \citep{es-etal-2024-ragas} has emerged and is very promising. Indeed, these methods offer a practical solution for generating datasets that mimic real human interactions by leveraging advanced LLM reasoning capabilities. Such synthetic datasets can include a broad range of scenarios, from straightforward factual questions to more nuanced domain-specific ones enabling robust and comprehensive evaluation of RAG systems. Additionally, synthetic data generation methods are increasingly recognized as vital components for the safe, transparent, and compliant evaluation of AI systems. In fact, regulatory frameworks, such as the European Union’s AI Act \citep{ai_act_2024}, explicitly promote the use of synthetic datasets within AI compliance and auditing processes. However, maintaining both a high efficiency of privacy-preserving mechanisms in retrieval systems and adherence to privacy regulations remains essential to building reliable and ethical evaluation frameworks.


In fact, according to the existing literature \citep{zeng2024good, ding2024survey}, retrieval systems may face serious privacy issues when the retrieval process involves sensitive data. For example, the authors in \citep{zeng2024good} observe that carefully designed user prompts are able to extract original sentences from the retrieval data or can also extract specific pieces of private information, potentially leading to the leakage of considerable amounts of retrieval data. The potential risk of information leakage can significantly limit the applications of retrieval systems. For example, a medical chatbot \citep{yunxiang2023chatdoctor} using patient history diagnosis cases as a source of knowledge can improve response quality but raises concerns about exposing sensitive patient information.

In this work, we take the first step toward exploring the generation of diverse and privacy-aware synthetic QA datasets designed specifically to serve as evaluation ground truth for assessing RAG systems. 
Our main contributions can be summarized as follows:

\begin{itemize}
    \item  
We introduce a modular and extensible multi-agent pipeline for synthetic QA dataset generation, designed specifically for evaluating RAG systems while ensuring a balance between diversity, privacy, and utility.

    \item 
We develop and perform a comprehensive twofold evaluation strategy: (1) diversity assessment combining qualitative judgments from LLM-based evaluators and quantitative diversity metrics, and (2) privacy assessment focused on the accuracy and effectiveness of entity masking across three specialized datasets.
\end{itemize}

\section{Related works}

\subsection{Retrieval-augmented generation and privacy issues}

Retrieval-Augmented Generation , introduced by \citep{lewis2020retrieval}, has gained substantial traction for enhancing LLM responses through external context. By retrieving relevant documents or passages and incorporating them into the prompt, RAG often yields output with improved accuracy and factual grounding \citep{gao2023retrieval}, mitigating the well-documented ``hallucination'' problem in LLMs \citep{shuster2021retrieval}. In addition to higher-quality outputs, RAG offers architectural flexibility by allowing independent upgrades to any of its components (e.g., data storage, retriever, or the core LLM) without requiring full model retraining \citep{shao2023enhancing,cheng2023lift}. These advantages have led to the adoption of RAG in diverse settings, from personal chatbots to highly specialized expert systems \citep{panagoulias2024augmenting}.

Despite its clear benefits, the retrieval process introduces privacy risks, particularly in domains handling sensitive user data. For instance, \citep{huang2023privacy} highlight privacy implications of retrieval-based language models, showing how training data and user inputs can be unintentionally exposed through retrieved passages. Other works have demonstrated that RAG models are susceptible to extraction attacks \citep{zeng2024good}, including membership inference and reconstruction attacks, which exploit learned representations to infer whether a user’s data was part of the training set \citep{qi2024follow}. Such vulnerabilities pose significant challenges for deploying RAG-based solutions in sensitive applications (e.g., healthcare, finance) where data privacy is paramount. 

\subsection{Synthetic data generation using Large Language Models}

Recent advances in LLMs have created significant interest in using them to automatically generate synthetic data. For instance, \citep{ye2022zerogen,meng2022generating} utilize zero-shot prompting to produce synthetic samples for tasks such as text classification and question answering, subsequently training smaller models on this generated data. \citep{gao2023self} introduce a noise-robust re-weighting framework to further refine data quality, while \citep{chen2023mixture} propose mixing a set of soft prompts and applying prompt tuning to enhance diversity in the generated text. Beyond prompt-based methods, \citep{yu2024large} examine specific attributes of the data itself, such as length and style, to diversify synthetic outputs. In parallel, a growing line of work has begun to address privacy concerns in synthetic data generation.

The authors in \citep{tang2023privacy} propose a few-shot approach, generating private in-context demonstrations backed by differential privacy guarantees, and \citep{xie2024differentially} design a private evolution algorithm that enforces differential privacy throughout the generation process. In \citep{mitigateprivacyrag}, the authors propose a novel
two-stage synthetic pipeline that includes attribute-based data generation, which aims to maintain key information, and iterative agent-based refinement, which further enhances the privacy of the input data in RAG systems.

\subsection{Synthetic question answer generation (QAG) and RAG evaluation}

A new wave of QAG and RAG evaluation approaches is redefining the field through synthetic data generation and dynamic assessment methods powered by LLMs. Rather than relying on static, human-curated datasets, recent efforts leverage LLMs to generate QA pairs and evaluate model outputs using automated scoring mechanisms, such as LLM-as-a-judge frameworks. Systems like RAGAS \citep{es-etal-2024-ragas} support scalable, domain-adaptable evaluation by conditioning synthetic QA generation on retrieved context and employing flexible, model-driven evaluation criteria. These methods offer the advantage of tailoring evaluation to specific domains and evolving data distributions. However, they also introduce new challenges, including maintaining content diversity, ensuring output consistency, and protecting sensitive information, particularly when operating over proprietary or privacy-sensitive corpora. Our work builds upon this emerging paradigm by incorporating explicit mechanisms to address these concerns, contributing a privacy and and diversity-aware framework for RAG evaluation.
\section{The proposed solution}
\label{sec:proposed_solution}

Algorithm~\ref{alg:proposed_solution} formally outlines the multi-agent procedure, detailing the sequential interaction between the diversity agent, privacy Agent, and QA curation agent. Given an input dataset $D$ and clustering hyperparameters, the process outputs a set of synthetic QA samples $D_{QA}$, enriched with semantic diversity and reinforced by privacy protections.

\begin{algorithm}[tbh!]
\caption{Multi-Agent Synthetic Evaluation Dataset Generation for RAG}
\label{alg:proposed_solution}
{%
Input: $D$ \& Original document \\
}
Output: {$D_{QA}$: A diverse, privacy-compliant synthetic QA dataset}

\textbf{Initialization:} $D_{\text{div}} \leftarrow \{\},\ D_{\text{priv}} \leftarrow \{\},\ D_{QA} \leftarrow \{\},\ Report_{\text{priv}}, Report_{QA} \leftarrow \emptyset$

\textbf{Stage 1: Diversity Agent}\;
\begin{enumerate}[label=\arabic*., leftmargin=4em]
    \item \textbf{Clustering:} Apply $k$-means clustering algorithm to group $D$ into $k$ clusters $\{C_1, C_2, ..., C_k\}$ based on text embeddings.
    \item \textbf{Representative Sampling:} For each cluster $C_i$, select a subset of representative samples $S_i$.
    \item \textbf{Aggregate Diverse Samples:} $D_{\text{div}} \leftarrow \bigcup_{i=1}^{k} S_i$
    
\end{enumerate}

\textbf{Stage 2: Privacy Agent}\;
\begin{enumerate}[label=\arabic*., leftmargin=4em]
    \item \textbf{For each} $S_i$ \textbf{in} $D_{\text{div}}$:
    \begin{itemize}
        \item Detect PII in each sample $x \in S_i$
        \item Pseudonymize each identified entity to produce $x'$
        \item Accumulate private samples: $D_{\text{priv}_i} \leftarrow D_{\text{priv}_i} \cup \{x'\}$

    \end{itemize}
        \item \textbf{Aggregate Private Documents:} $D_{\text{priv}} \leftarrow \bigcup_{i=1}^{k} D_{\text{priv}_i}$

    \item \textbf{Privacy Report:} Record types and frequencies of pseudonymized entities in $Report_{\text{priv}}$
\end{enumerate}

\textbf{Stage 3: QA Curation Agent}\;
\begin{enumerate}[label=\arabic*., leftmargin=4em]
    \item \textbf{For each} $x' \in D_{\text{priv}}$:
    \begin{itemize}
        \item Generate $n$ QAs pair $(q, a)$
        \item Accumulate: $D_{QA} \leftarrow D_{QA} \cup \{(q, a)\}$
    \end{itemize}
    \item \textbf{Generate QA Report:}  Log model settings, number of successful QA pairs, failures and generation procedure in $Report_{QA}$
\end{enumerate}

\Return $D_{QA},\ Report_{\text{priv}},\ Report_{QA}$

\end{algorithm}

In the first step, the Diversity agent clusters the original dataset using semantic embeddings and selects representative samples from each cluster, ensuring a broad coverage of topics. The second step, the privacy agent, operates over each cluster’s representative samples, detecting and pseudonymizing sensitive entities to produce a private version of the data along with a structured privacy report. Finally, the QA curation agent synthesizes question-answer pairs from the private data, generating evaluation-ready samples along with a QA generation report that summarizes success rates and generation dynamics.
\section{Experiments}
\subsection{Experimental Setup}
\label{sec:experimental_setup}

\paragraph{Implementation details}
All components of our multi-agent framework were implemented in Python, using the LangGraph framework to orchestrate inter-agent communication and control flow. Each agent was instantiated as a node within a LangGraph workflow.

\paragraph{Language models}
We employed models from Azure OpenAI services. Specifically, we used \texttt{GPT-4o} for the Diversity agent and the QA curation agent, due to its fast response time and strong generalization capabilities for content generation. For the privacy agent, we used \texttt{GPT-4.1}, which offers superior reasoning and tool-usage capabilities that are crucial for accurate PII detection and transformation tasks involving interaction with APIs or complex instructions. To ensure reproducibility and minimize variability in outputs, the temperature of all language models was fixed at $0$ during inference.
For the clustering process in the diversity agent, we generated embeddings using OpenAI’s \texttt{text-embedding-3-small} (Ada 3) model with an embedding dimension of $1536$. Input documents were preprocessed into chunks of $256$ tokens before applying $k$-means clustering.

\paragraph{Agents tool configuration}
Each agent in the system operates with tailored tools suited to its objectives. First, the diversity agent uses a $k$-means clustering function to identify latent topic clusters within the input document. The optimal value of $k$ is selected using intra-cluster distance scores.

Second, the privacy agent performs pseudonymization based on a predefined set of PII categories. It scans the generated content, identifies sensitive entities, and replaces them using context-aware transformations. In addition, it produces a structured privacy report detailing which PIIs were correctly detected, masked, or missed.

Last, the QA curator agent generates final QA pairs from the enriched, privacy-preserved inputs by leveraging advanced prompting techniques. It also produces a comprehensive generation report summarizing the types of QA pairs created, their alignment with source content, and overall dataset characteristics.

\subsection{Research question 1: How does the proposed multi-agent system enhance the diversity of the generated evaluation dataset?}
\label{sec:rq1_diversity}

\subsubsection{Baselines}

To assess the effectiveness of our proposed multi-agent approach in enhancing dataset diversity, we compare against two baselines:

\paragraph{(1)} Evolutionary generation (RagasGen).
Inspired by works such as Evol-Instruct and RAGAS \citep{es-etal-2024-ragas}, this baseline uses an evolutionary generation paradigm to produce QA pairs. It iteratively mutates and refines questions to maximize diversity along dimensions such as reasoning complexity, multi-hop dependencies, and topic breadth. \paragraph{(2)} Direct Prompting (DirPmpt).
This baseline uses direct LLM prompting with few shot examples. A GPT-4o model is prompted with handcrafted instructions to produce diverse QA pairs.
\subsubsection{Diversity evaluation dataset}

To evaluate our multi-agent framework's ability to generate diverse QA pairs, we use the official EU AI Act as input. Its rich structure and varied content provide a realistic and challenging testbed for assessing diversity in synthetic evaluation sets.

\subsubsection{Diversity evaluation methodology}

To assess the diversity of the generated QA sets, we use the LLM-as-a-Judge approach, where GPT-4.1 is prompted to act as an expert evaluator. The model receives pairs of evaluation sets, our generated set and baseline sets, along with instructions to judge question diversity based on semantic variety, topical coverage, and phrasing differences. It then assigns diversity scores on a scale from 1 to 10. Additionally, we use the 
\textbf{CosineSimilaritytoDiversity}\citep{cosine}, which inverts the average pairwise cosine similarity of sentence embeddings, lower values indicate greater semantic spread.
\subsubsection{Findings discussion}

First, we observe that our multi-agent system outperforms RagasGen and DirPrmpt in all evaluated settings, with consistent gains observed across both 
qualitative and quantitative metrics.

Furthermore, we observe a consistent trend across all diversity measures: as the test set size increases, so does the diversity of the generated questions. The LLM-as-a-Judge scores (GPT-4.1) rise from 7.8 at 10 samples to 9 at 100 samples, indicating that the generated question sets increasingly exhibit richer topic coverage and variation in structure. Quantitatively, the \texttt{CosineSim2toDiversity} score becomes less negative (closer to zero), reflecting that questions are increasingly dissimilar to each other, a direct proxy for higher diversity. These results demonstrate that our multi-agent system enhances question diversity, particularly at larger scales.

\begin{table}[tbh!]
\label{tab:diversity_results}
\centering
\begin{tabular}{c|ccc|ccc}
\toprule
QA set size & \multicolumn{3}{c|}{GPT-4.1 Diversity Rating} & \multicolumn{3}{c}{Cosine Sim. to Diversity} \\
& Ours & RagasGen & DirPmpt & Ours & RagasGen & DirPmpt \\
\midrule
10 & \textbf{7.8} & 7.0 & 6.2 & \textbf{-0.36} & -0.40 & -0.45 \\
25 & \textbf{8.2} & 7.3 & 6.3 & \textbf{-0.31} & -0.38 & -0.43 \\
50 & \textbf{8.6} & 7.4 & 6.9 & \textbf{-0.26} & -0.36 & -0.38 \\
75 & \textbf{8.9} & 8.0 & 7.5 & \textbf{-0.18} & -0.34 & -0.35 \\
100 & \textbf{9.0} & 8.1 & 7.6 & \textbf{-0.15} & -0.33 & -0.33 \\
\bottomrule
\end{tabular}
\caption{Diversity and similarity metrics comparison between question sets generated by our method, RagasGen, and DirPmpt.}
\end{table}

\subsection{Research Question 2: How does the proposed multi-agent solution preserve the overall privacy of the system?}

\subsubsection{Privacy evaluation datasets}

To evaluate the effectiveness of the privacy agent, we use three benchmark datasets provided by AI4Privacy\footnote{\url{https://huggingface.co/ai4privacy}}: \textbf{PII-Masking-200K}, \textbf{PWI-Masking-200K}, and \textbf{PHI-Masking-200K}. These tabular datasets contain long-form sentences annotated with private entities from different domains. The PWI dataset includes job titles, company names, and salary information. The PHI dataset focuses on medical diagnoses, genetic information, and gender. The PII dataset contains names, locations, dates of birth, and contact details. Each dataset also provides additional metadata such as entity type, position, and frequency.

To simulate realistic input conditions for our pipeline, we concatenated individual samples from each dataset into longer text paragraphs. We refer to each resulting dataset as PWI, PHI, or PII. Each consists of domain-specific long sentences containing private entities and their corresponding masked versions. Table 2 summarizes the main statistics for each dataset, including document length and the number of private entities.

\begin{table}[h]
\centering
\begin{tabular}{l|c|c|c}
\toprule
\textbf{Dataset} & \textbf{Dataset length (sentences)} & \textbf{Total entities number} & \textbf{Avg entities per sentence} \\
\midrule
PWI  & 1800 & 451 & 3.99 \\
PHI  & 1700 & 422 & 4.02 \\
PII  & 1600 & 591 & 2.71 \\
\bottomrule
\end{tabular}
\caption{Statistics of the constructed privacy evaluation datasets.}
\label{tab:privacy_datasets}
\end{table}

\subsubsection{Experimental results}

\begin{figure}[h]
\centering
\includegraphics[width=0.8\textwidth]{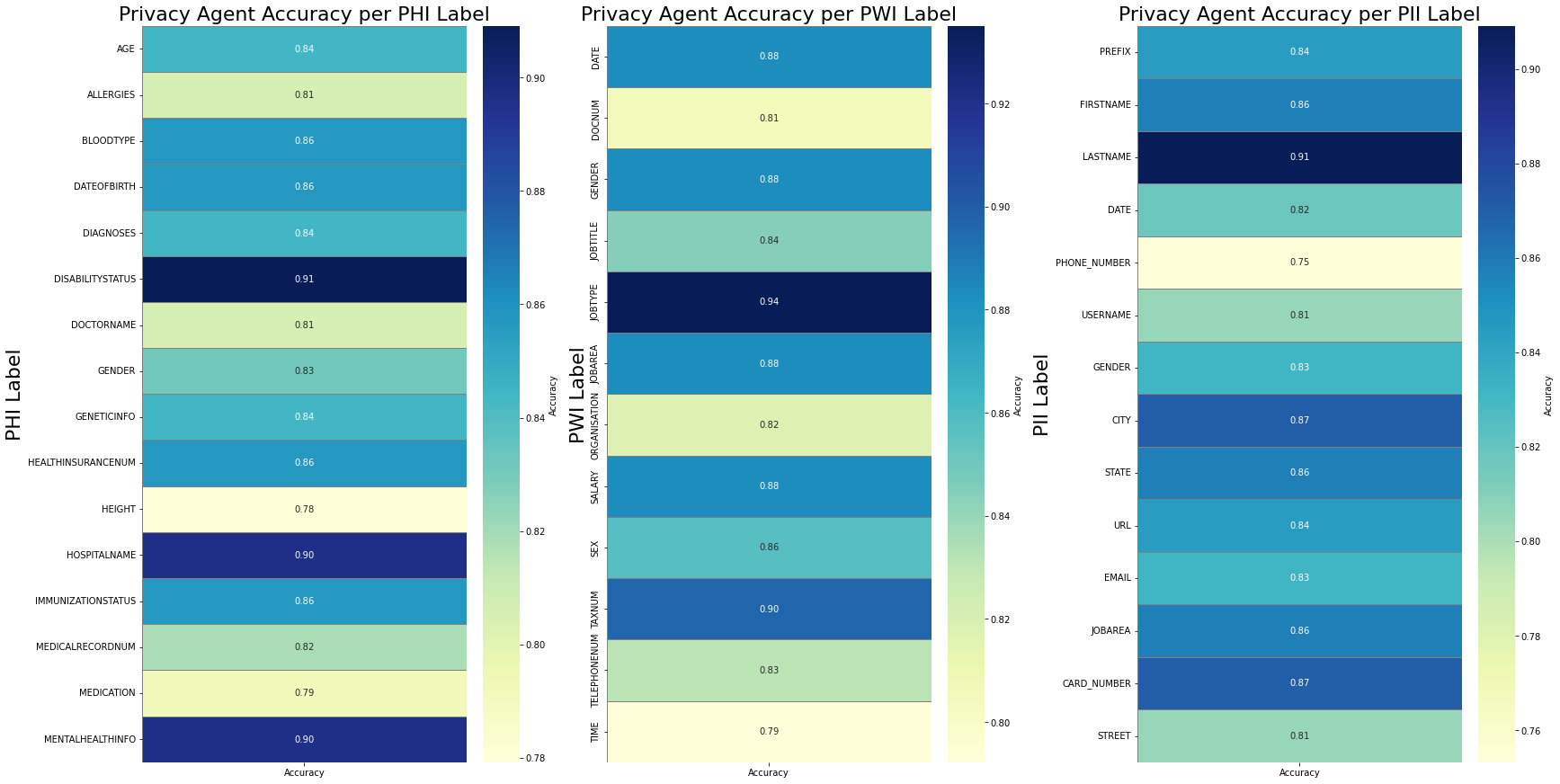}
\caption{Privacy agent accuracy per entity type across the PHI, PWI, and PII datasets.}
\label{fig:privacy_accuracy}
\end{figure}

In Figure~\ref{fig:privacy_accuracy} we present label-wise accuracy across the PHI, PWI, and PII datasets The privacy agent shows strong overall performance, with most labels achieving accuracies between 0.75 and 0.90. On the PHI dataset, the highest scores are observed for DISABILITYSTATUS (0.91), HOSPITALNAME (0.90), and MENTALHEALTHINFO (0.90), indicating robust handling of sensitive medical information. On the PWI dataset, the model performs best on JOBTYPE (0.94), TELEPHONENUM (0.90), and DATE, GENDER, SALARY, ORGANISATION, DBAREA (all 0.88), demonstrating high reliability in identifying entities related to the workplace. Moreover, results on the PII dataset highlight strong performance for LASTNAME (0.91), CARDNUMBER and CITY (0.87), and FIRSTNAME, STATE, and JOBAREA (all 0.86), confirming the agent's effectiveness in detecting general personal identifiers.

Interestingly, overlapping labels such as GENDER appear across PHI (0.83), PWI (0.88), and PII (0.83), and yield consistently strong scores. This suggests that the privacy agent generalizes well across domains.

\section{Conclusion and future work}

In this work, we introduced a modular multi-agent framework for the generation of synthetic QA datasets tailored to the rigorous evaluation of RAG systems. Our approach decomposes the dataset construction process into distinct, specialized agents, each focused on enriching semantic diversity, enforcing privacy safeguards, and curating high-quality QA pairs. Through comprehensive experiments we highlight the framework’s effectiveness in producing evaluation datasets that are both representative and privacy-preserving, addressing critical challenges in real-world RAG evaluation.

Looking ahead, we aim to enhance the autonomy and collaboration of individual agents by leveraging tool-augmented foundation models. For example, the diversity agent could dynamically infer optimal clustering structures, while the privacy agent could adaptively identify and transform PIIs beyond static entity lists. In addition, we plan to explore agent-to-agent communication protocols and effective independent agent collaboration, potentially through frameworks like model context protocol, to improve coordination and task delegation among agents.

Future work will also include rigorous evaluation of the framework’s resilience to privacy attacks, helping to clarify its defensive boundaries and inform improvements. As AI regulations such as the EU AI Act continue to evolve, subsequent versions of our framework will further align synthetic evaluation set generation not only with principles of technical trustworthiness, but also with emerging regulatory requirements.

\bibliography{sample-ceur}

\end{document}